\documentclass[conference]{IEEEtran}
\IEEEoverridecommandlockouts
\usepackage{cite}
\usepackage{amsmath,amssymb,amsfonts}
\usepackage{algorithmic}
\usepackage{graphicx}
\usepackage{textcomp}
\usepackage{xcolor}
\usepackage{float} 
\usepackage{xcolor}
\usepackage{tabularx}
\def\BibTeX{{\rm B\kern-.05em{\sc i\kern-.025em b}\kern-.08em
    T\kern-.1667em\lower.7ex\hbox{E}\kern-.125emX}}
    
\begin{document}

\title{Hybrid Efficient Unsupervised Anomaly Detection for Early Pandemic Case Identification\\
}
\author{\IEEEauthorblockN{Ghazal Ghajari}
\IEEEauthorblockA{\textit{Computer Science and Engineering } \\
\textit{Wright State University}\\
Ohio, USA \\
ghajari.2@wright.edu}
\and
\IEEEauthorblockN{Mithun Kumar PK}
\IEEEauthorblockA{\textit{Computer Science and Engineering} \\
\textit{Wright State University}\\
Ohio, USA \\
pk.2@wright.edu}
\and
\IEEEauthorblockN{Fathi Amsaad}
\IEEEauthorblockA{\textit{Computer Science and Engineering} \\
\textit{Wright State University}\\
Ohio, USA \\
fathi.amsaad@wright.edu}
}

\maketitle

\begin{abstract}
Unsupervised anomaly detection is a promising technique for identifying unusual patterns in data without the need for labeled training examples. This approach is particularly valuable for early case detection in epidemic management, especially when early-stage data are scarce. This research introduces a novel hybrid method for anomaly detection that combines distance and density measures, enhancing its applicability across various infectious diseases. Our method is especially relevant in pandemic situations, as demonstrated during the COVID-19 crisis, where traditional supervised classification methods fall short due to limited data. The efficacy of our method is evaluated using COVID-19 chest X-ray data, where it significantly outperforms established unsupervised techniques. It achieves an average AUC of 77.43\%, surpassing the AUC of Isolation Forest at 73.66\% and KNN at 52.93\%. These results highlight the potential of our hybrid anomaly detection method to improve early detection capabilities in diverse epidemic scenarios, thereby facilitating more effective and timely responses.
\end{abstract}


\section{Introduction}
Machine learning is extensively utilized in healthcare applications, including preventive measures like detecting organic volatile compounds as carcinogenic elements in the atmosphere \cite{a21}, or identifying diseases early during epidemics, and for diagnostic purposes. Early detection is crucial for controlling epidemics and implementing timely preventive measures. However, the limited availability of case data during the initial stages of an outbreak poses significant challenges for training machine learning models \cite{a1}. This issue was particularly evident during the COVID-19 pandemic, where data scarcity at the initial stages hindered the prompt implementation of preventive actions, leading to devastating consequences \cite{a2}. The first reports of COVID-19 emerged in late 2019 \cite{a3}, but the initial lack of comprehensive data delayed widespread recognition and response, contributing to the tragic loss of millions of lives. The subsequent development and distribution of effective vaccines have shown potential ways to mitigate such crises \cite{a4}. 

The lessons learned from the COVID-19 pandemic emphasize the importance of early case detection in preventing future health disasters. This paper uses COVID-19 as a case study to analyze the illness resulted from the SARS-CoV-2 virus \cite{a5}. While the primary diagnostic approach is reverse transcription-polymerase chain reaction (RT-PCR), it is both costly and time-intensive \cite{a6}. Research demonstrates the potential of imaging methods such as chest X-rays (CXR) and computed tomography (CT) scans, in identifying COVID-19 cases early, despite their variability and inherent challenges \cite{a7,a8, aNew1}. Most studies using CT scans and CXR for COVID-19 diagnosis rely on supervised classification strategies, which require extensive data that may not be available in the early stages of an outbreak \cite{a9, aNew2}. This dependence on large datasets, only obtainable after numerous infections, is both unethical and impractical for early epidemic detection. 

This paper presents a novel hybrid unsupervised anomaly detection method that integrates distance and density measures for early pandemic case identification. This approach is particularly effective in scenarios with limited early-stage data, offering a versatile solution applicable across various infectious diseases. By combining these measures, our method enhances the detection of anomalies without the need for extensive training data\cite{a10}, providing a robust tool for early epidemic detection and broadening the scope of anomaly detection in other domains such as network security and financial data analysis \cite{a11}.

\section{Related Works}

In the domain of unsupervised anomaly detection, comparative studies are scarce. New algorithms are often benchmarked against advanced techniques like the local outlier factor (LOF)\cite{b12} and k-nearest neighbors (KNN)\cite{b13}, however these comparisons are frequently limited by the availability of datasets and a lack of uniform evaluation standards \cite{a12}, also Studies tend to focus on specific applications, often using single datasets, which may not generalize across different scenarios. For example, various anomaly detection methods have been applied to distinct fields such as intrusion detection, space shuttle engine failures, and marine video surveillance, using datasets like KDD-Cup99 or proprietary data that are not publicly accessible \cite{a13}.

Unsupervised anomaly detection can be categorized into methods based on nearest neighbors, clustering, statistical approaches, and more recently, subspace techniques. Each category has its strengths and is suited to particular kinds of data challenges \cite{a14}. By leveraging unsupervised anomaly detection and a novel hybrid model, we aim to enhance early epidemic detection. Our approach not only addresses the gaps in future epidemic response strategies but it also contributes to the broader field of anomaly detection across various domains \cite{a15}.

In this study, we introduce a novel hybrid model that integrates the concepts of distance and local density for anomaly detection. This model is versatile, accommodating various types of data features, including numerical and nominal. We contend that this method offers a more holistic and complete approach to addressing the challenges of anomaly detection. Its unsupervised nature ensures that it does not rely on potentially inaccurate assumptions about data structure, and it remains effective even in the absence of extensive training data. Moreover, we employ principles from graph theory to innovatively calculate and define local density, enhancing our model's ability to identify anomalies. This approach not only enriches our understanding of graph-based techniques but also expands the possibilities for their application in detecting anomalous patterns.

\section{Methodology}
\begin{figure*}[!t]
    \centering
    \includegraphics[width=\textwidth]{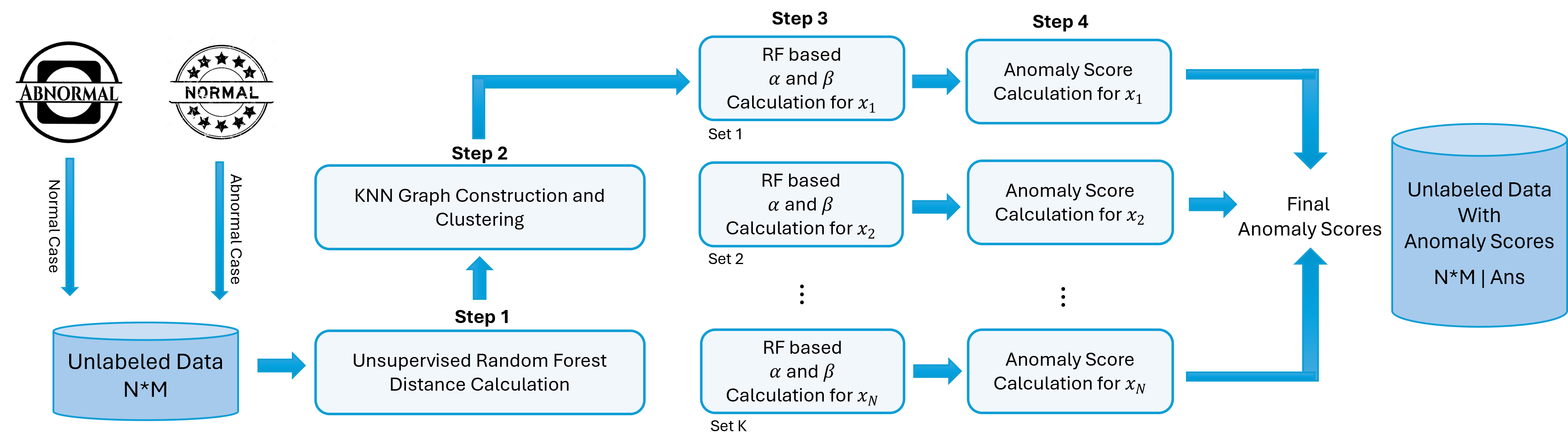}
    \caption{The flowchart of unsupervised hybrid proposed algorithm.}
    \label{fig}
\end{figure*}

In this section, a new method is presented for the anomaly detection challenge. Fig.\ref{fig} illustrates the general procedure of the proposed algorithm. In brief, the process begins with the input data being processed by the random forest algorithm, which produces a complete distance (or dissimilarity) matrix for all data elements. Using this distance matrix, a k-nearest neighbor (KNN) graph is constructed to model the data manifold. This KNN graph is then clustered using a graph community detection algorithm, which automatically divides the data into several clusters. The center of each cluster is used as the representative of that cluster in the next step. Two parameters, density and distance, are calculated for each data point. Finally, the anomaly criterion is determined by combining these density and distance parameters for each data point. From the calculated (and normalized) anomaly scores, abnormal data can be easily distinguished from normal data. The following sections elaborate on the steps of the proposed algorithm.\\

\noindent \textbf{Step 1: }Calculation of distance matrix by unsupervised random forest which is motivated by three main reasons:

\begin{enumerate}
\item  Random Forest considers the operational contribution of each data feature on the calculated distance while considering the relationship and correlation between features in this calculation.
\item  The tree structure in the decision trees used in the random forest is able to model the data with a hierarchical manifold in a characteristic form.
\item  Random forest works equally well with different types of attributes, including numerical, nominal, or group, with no need to change and program the data.
\end{enumerate}

A random forest consists of a set of decision trees. The random forest method is a supervised machine learning approach extensively employed for classification and regression tasks. In this research, we intend to use its unsupervised form previously presented in a study to calculate a complete distance matrix \cite{a16}. 

The main idea of using a random forest structure is to produce a distance matrix in which feature ranks and inter feature dependencies are used to distinguish original data $X$ from fake data $Y$. The dummy data $Y$ is created by sampling the product of the marginal distributions of the features in $X$. This means that for each selected data characteristic, its marginal distribution is calculated, then random sampling is done from this distribution to the number of $X$ points. This operation is repeated for all $X$ features and $Y$ data is produced by putting these features produced by sampling together. 

For simplicity, we label the original data $X$ with class 1 and the dummy data $Y$ with class 2. Then we create the $XY$ dataset from the set of both $X$ and $Y$ data and input it into the random forest. We represent the entire set of data features by $F = \{ f_1, f_2, \ldots, f_m \}$. By employing random forest classification, the model aims to differentiate between data $X$ and randomly generated data $Y$.

The random forest algorithm causes the creation of several sets of decision trees in which various data points from $X$ are distributed across different branches of these trees. This distribution is based on the size and the degree of correlation between the features of $X$ (set $F$). The process ensures that the closer two data points are to each other, the smaller the average tree distance between them. We use the concept of entropy to create decision trees. Entropy for the subset $\psi$ of the integration data set (class 1 and class 2) is defined by formula 1:
\begin{equation}
E_{fi}(\Psi) = -p_{1.i} \log_2 p_{1.i} + p_{2.i} \log_2 p_{2.i} \label{eq}
\end{equation}
Where $p_{1.i}$ and $p_{2.i}$ are the shares of classes 1 and 2 in $\psi$ when the feature $f_i$ is used to divide $\psi$. Similarly, for each feature $f_i$ in a particular node of the decision tree, $IG$ (expected decrease in entropy) can be calculated in formula 2: 
\begin{equation}
IG(\Psi.f_i)=E_{f_i}(\Psi)-\sum_{\phi \in \text{values}(f_i)} \frac{|\Psi_\phi|}{|\Psi|} E_{f_i} (\Psi_\phi) \label{eq}
\end{equation}
Where $\psi_\phi$ is the subset of $\psi$ with the value $\phi$ for the feature $f_i$. Finally, the attribute $f_i$ with the highest $IG$ value is selected to divide the data at that node. When the whole tree is formed using information gain $(IG)$ and entropy concepts, each leaf node contains one or more data. In short, the degree of similarity of the random forest between two data $x_i$ and $x_j$ can be defined as follows: the number of common nodes between $x_i$ and $x_j$ that exist in the root path of the tree $\theta$ to each of them, which is shown $S_{RF}(x_i,x_j)$ suppose that $T$ trees are randomly generated in the forest. The final distance of the random forest between $x_i$ and $x_j$ is calculated by formula 3:
\begin{equation}
D_{RF}(x_i, x_j) = \sqrt{1 - \sum_{\theta=1}^T \frac{S_{RF}(x_i, x_j)}{H_\theta \times T}} \label{eq}
\end{equation}
Where $H_\theta$ represents the height of the tree $\theta$ and $H_\theta \times T$ is used as a factor for normalization in the denominator of the fraction. Using this technique, the complete distance matrix for $X$ points is calculated, which we call $D_{RF}$. \\

\noindent \textbf{Step 2: }To construct a KNN (k-nearest neighbors) graph from the calculated distance matrix $D_{RF}$ , we first identify the k smallest distances for each node, excluding the node's distance to itself. These $k$ smallest distances represent the $k$ nearest neighbors of each node. We then create an undirected graph where each node is connected to its $k$ nearest neighbors, with edges representing the nearest neighbor relationships based on the distances in $D_{RF}$ . This is achieved by iterating through each node, selecting its  $k$ nearest neighbors, and adding the corresponding edges to the graph. In this work, $k$ is experimentally set to $log(N)$.\\

\noindent \textbf{Step 3:} Calculation of density parameters and anomaly distance. As discussed, in the second step of the proposed clustering algorithm, our data was automatically divided into k distinct clusters. Then, from each cluster, its center was chosen as a representative. In this step, using the selected points, we calculate the two parameters of density and distance for all data points.

Density parameter: Using the distance matrix calculated by the random forest method and the graph formed for clustering in the second step, for each data we define the density parameter $\alpha_i$ as formula 4:
\begin{equation}
\alpha_i =\sum_j f(d_{ij}-d_c) \hspace{0.5cm}
f(x) = \begin{cases} f(x)=1 & x < 0 \\f(x)=0 & \text{otherwise} \end{cases}
 \label{eq}
\end{equation}
$d_c$ is a cluster specific threshold value and $d_{ij}$ is tree-based distance between points $i$ and $j$. Briefly, this means the number of points that are closer than dc to the point $i$ in related specific cluster. In our work, for each cluster $C_k$, define dc based on the distribution of the pairwise distances within the cluster. We use a percentile of the distance distribution to ensure that dc is representative of the local density characteristics.

Distance parameter: Once we have calculated the density for all the data, we calculate the distance parameter by calculating the tree distance to $M$ the center of the denser cluster using formula 5:
\begin{equation}
\beta_i = \frac{ \sum_{j: \alpha_j > \alpha_i, j \in \Psi} D_{RF}(x_i,x_j)} {M} \label{eq}
\end{equation}

\noindent \textbf{Step 4:} When we calculate the two parameters of density $\alpha$ and distance $\beta$ for all data points, we obtain a norm/abnormality score for each data using formula 6:
\begin{equation}
\text{A}_i = \frac{\beta_i}{\alpha_i} \label{eq}
\end{equation}
It is clear that the value of $A_{i}$ is larger for abnormal data because these data are thinner, that is, they have a lower neighborhood density, but on average, they are located at farther distances from the high-density points (centers of clusters). This scenario is wholly opposite to norm data. This means
that the norm data has high density and low distance parameters. Therefore, $A_{i}$ can be reliably used as a criterion to distinguish between normal data and abnormal data. 

However we need to define a suitable threshold to distinguish normal from abnormal data points within calculated $A_i$ scores. To define this threshold, which in real scenarios, should exhibit a highly skewed distribution, we apply a log transformation to reduce skewness, making the distribution more symmetrical. Specifically, we transform the scores using $A_{i}^{'}=log (A_i+1)$, where adding 1 ensures that zero scores are handled appropriately. We then calculate the mean ($\mu A_{i}^{'}$) and standard deviation ($\sigma A_{i}^{'}$) of the transformed scores. Using the z-score method, we set a threshold at $\mu A_{i}^{'}+z.\sigma A_{i}^{'}$ with $z$ typically chosen between 2.5 and 3 for robust outlier detection. Finally, we convert this threshold back to the original scale with $Threshold~A_{i}$=$exp (Threshold~A_{i}^{'}) - 1$. This method effectively handles skewed distributions and ensures robust identification of abnormal data points.

\begin{figure*}[!t]
    \centering
    \includegraphics[scale=0.7]{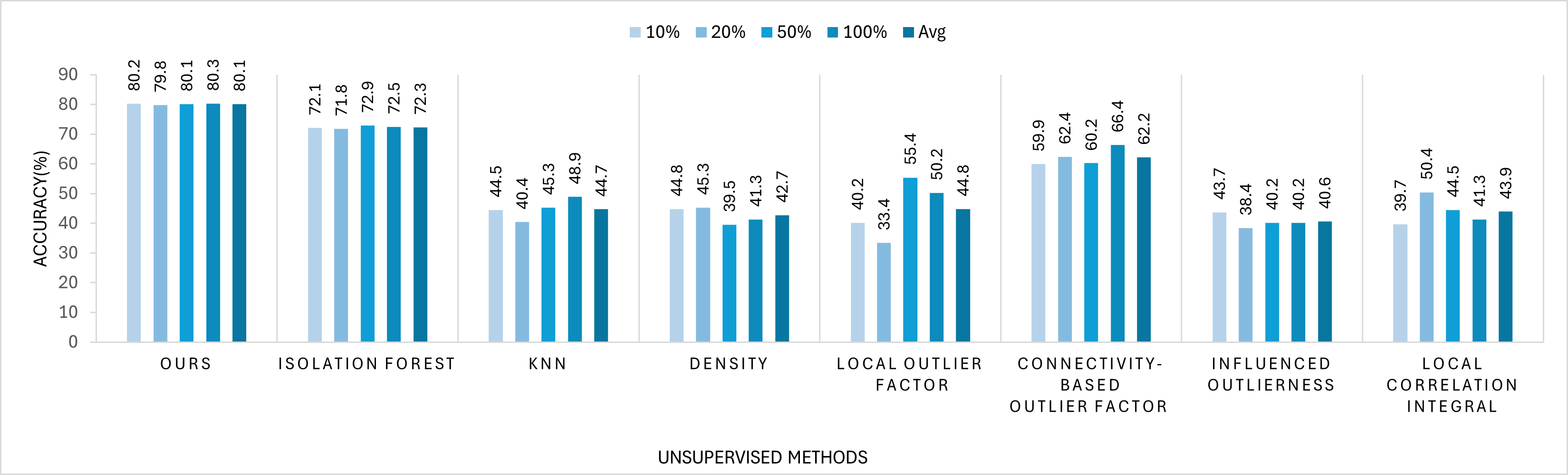}
    \caption{Stability Evaluation Considering Different Methods}
    \label{fig2}
\end{figure*}
\begin{figure*}[t]
    \centering
    \includegraphics[scale=0.7]{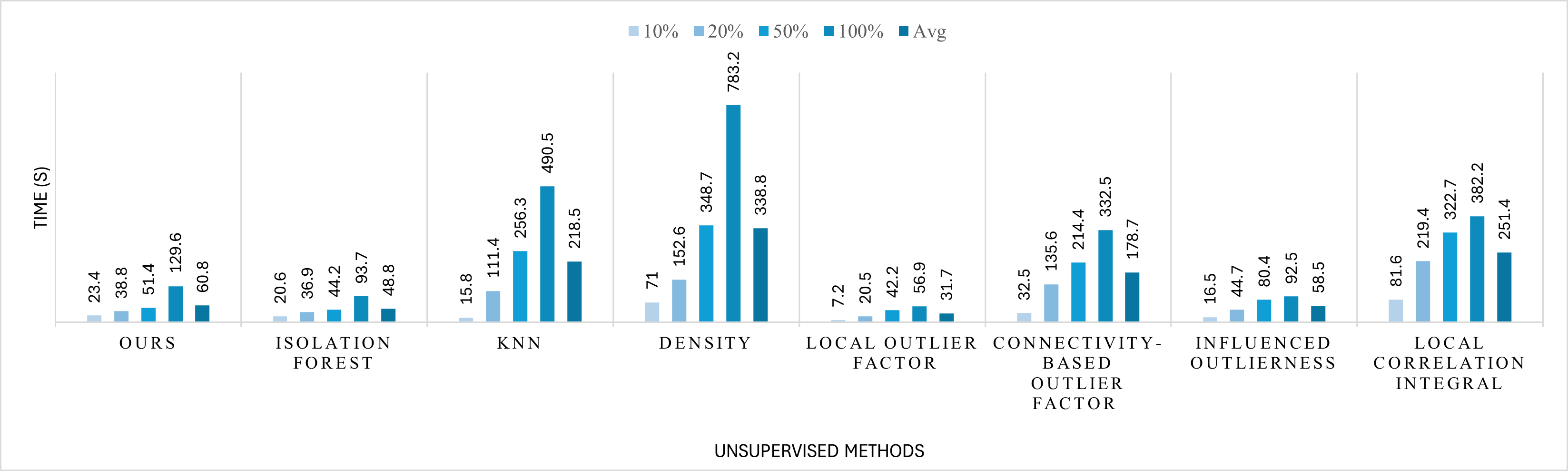}
    \caption{Evaluation of Run-time Considering Different Methods}
    \label{fig3}
\end{figure*}
\section{Experimental Results}
To evaluate our method for identifying early-stage pandemic cases, we utilized the COVIDx dataset \cite{a19}, a continually updated project with a variable sample size. This dataset includes 8,066 chest x-ray (CXR) images from ``healthy" individuals and 358 CXR images from ``COVID-19" patients. For this study, the dataset comprises chest x-ray images used with three pre-trained deep learning models: VGG-19 \cite{a17}, ResNet-50 \cite{a18}, and COVID-Net \cite{a19}, to implement the method and extract numerical features from each image.

Our evaluation metric in all experiments is the Area Under the Curve (AUC)\cite{a20}. The AUC assesses a model's capability to differentiate between classes (normal versus abnormal) and summarizes the receiver operating characteristic (ROC) curve. The AUC score varies between 0 and 1, with higher scores indicating superior model performance. In this study, we present the AUC values as percentages to aid interpretation. For example, an AUC of 0.78 is reported as 78\%.

Table \ref{tab1} details three supervised deep learning methods that achieve superior results due to pre-training, although they require substantial training data. All these results come from \cite{a19}. In real-world applications, a significant limitation is the lack of sufficiently large datasets for training. As a result, unsupervised methods are recommended. Additionally, deep learning methods with extensive networks encounter challenges in parameter management and model stability in response to minor data alterations. Our proposed method, which utilizes feature extraction with COVID-Net, addresses these issues effectively.

\begin{table}
\caption{Performance Evaluation Considering Different Architectures}
\centering 
\begin{tabular}{|l|l|l|l|}\hline
Architecture\textcolor{white}{This textisred}& Params (M)& MACs (G)& Acc. (\%) \\ \hline
VGG-19& 20.37 &89.63& 83.2\\ \hline
ResNet-50& 24.97& 17.75& 90.6\\ \hline
COVID-Net &11.75& 7.5 &93.3\\ \hline
Ours& - &0& 80.3\\ \hline
\end{tabular}
\label{tab1}
\end{table}

In the next experiment, we conduct feature extraction using three deep learning models—VGG-19, ResNet-50, and COVID-Net—to extract features from each image in COVID-X dataset. Subsequently, we apply various unsupervised anomaly detection methods to evaluate their performance in detecting COVID-19 cases (anomalous cases) versus normal ones. As shown in Table \ref{tab2}, our algorithm outperforms other unsupervised methods in terms of accuracy.

\begin{table}
\caption{AUC Evaluation Considering Different Methods}
\begin{center}
\begin{tabular}{|l|l|l|l|l|}
\hline
Method &VGG-19 &\begin{tabular}[c]{@{}l@{}}ResNet\\-50\end{tabular}& \begin{tabular}[c]{@{}l@{}}COVID\\-Net \end{tabular}& Avg\\ \hline
Ours &74.2 &77.8 &80.3 &77.4\\ \hline
Isolation Forest& 76.3& 72.2 &72.5 &73.6\\ \hline
Knn& 55.3 &54.6 &48.9& 52.9\\ \hline
Density &49.4& 50.2& 41.3 &46.9\\ \hline
Local Outlier Factor& 54.3 &55.3 &50.2 &53.2\\ \hline
\begin{tabular}[c]{@{}l@{}}Connectivity-Based \\Outlier Factor\end{tabular}& 43.2 &40.3 &66.4& 49.9\\ \hline
Influenced Outlierness& 36.4 &49.8 &40.2& 42.1\\ \hline
Local Correlation Integral &49.1 &52.3& 41.3 &47.5\\ \hline
\end{tabular}
\label{tab2}
\end{center}
\end{table}

Fig. \ref{fig2} illustrates the stability of different unsupervised methods by testing their performance across various dataset sizes. In this scenario, subsamples of the COVID-X dataset (10\%, 20\%, 50\%, and 100\%) are used. After feature extraction with COVID-Net, unsupervised methods are applied to these subsamples. The figure reports the average accuracy, demonstrating that our proposed technique maintains more stable performance compared to other methods.

Fig. \ref{fig3} compares the execution times of different algorithms. While the Local Outlier Factor (LOF) algorithm is the fastest, it lacks accuracy and stability. The Isolation Forest (IF) algorithm strikes a balance between speed and accuracy. However, our proposed method excels, outperforming others in terms of both accuracy and overall average performance. We attribute this superior performance to the hybrid nature of our approach, which simultaneously utilizes the concepts of local density and distance to the cluster centroid.

Testing the algorithm with various subsets of data, such as 10\%, 20\%, and 50\%, as depicted in Fig. \ref{fig2} and Fig. \ref{fig3}, demonstrates the robustness and versatility of the proposed method. This approach allows for an evaluation of the algorithm's performance under different data availability scenarios, which is crucial for early-stage pandemic detection where data might be scarce. By maintaining high accuracy and stability even with smaller subsets, the algorithm proves its efficacy in real-world applications where data collection is often incremental. This also highlights the method's scalability and its ability to provide reliable results despite limited initial data, ensuring that early detection and timely intervention can be achieved.

\section{Conclusions and Future Work}
In this study, a novel unsupervised hybrid method is proposed that utilizes both distance and density measures to identify early cases of epidemics, such as COVID-19. This method offers several benefits over previous approaches, including its robustness, faster processing, and the elimination of the need for training; it is also capable of identifying anomalies within predominantly small clusters. Despite COVID-19's progression to pandemic status, we contend that anomaly detection models can reliably spot early instances in future pandemics or new waves of current ones using only typical chest x-ray (CXR) images found in various medical settings. Looking ahead, we plan to adapt and enhance this algorithm for use in natural language processing (NLP) and to incorporate ideas from weak supervision to boost our algorithm's effectiveness. We believe that incorporating weak supervision will markedly enhance the method's efficiency. Additionally, this algorithm can be modified to run in parallel, further enhancing its speed.

\section{Further Discussion on Potential Applications}

The hybrid unsupervised anomaly detection method introduced in this study has significant potential applications beyond pandemic case identification, which can further strengthen its conclusions. For instance, in the healthcare sector, this method can be employed to detect anomalies in patient data for early diagnosis of various diseases, thereby improving patient outcomes. In network security, it can identify unusual patterns of network traffic that indicate potential cyber threats, enhancing the security of digital infrastructures. Additionally, financial institutions can use this method to detect fraudulent transactions early, minimizing financial losses. The algorithm can also be adapted for predictive maintenance in industrial settings, identifying early signs of equipment failure from sensor data to prevent costly downtimes. Finally, environmental monitoring can benefit from this approach by detecting anomalies in climate data, aiding in the timely response to environmental changes and natural disasters. These diverse applications underscore the algorithm's versatility and robustness, making it a valuable tool across multiple domains.


\end{document}